\documentclass[11pt]{article}
\usepackage{coling2020}
\usepackage{times}
\usepackage{url}
\usepackage[utf8]{inputenc}
\usepackage{dirtytalk}
\usepackage{latexsym}
\usepackage{hyperref}
\usepackage{graphicx}
\usepackage{epstopdf}
\usepackage{caption}
\usepackage{subcaption}
\usepackage{amssymb}
\usepackage{amsmath}
\usepackage{hyperref}
\usepackage{appendix}
\colingfinalcopy % Uncomment this line for the final submission

\title{PoinT-5: Pointer Network and T-5 based Financial Narrative Summarisation}

\author{Abhishek Singh \\
  Samsung R\&D Bangalore \\

  {\tt abhishek.s.eee15@iitbhu.ac.in} }

\date{}

\begin{document}
\maketitle
\begin{abstract}
  Companies provide annual reports to their shareholders at the end of the financial year that describes their operations and financial conditions. The average length of these reports is 80, and it may extend up to 250 pages long. In this paper, we propose our methodology PoinT-5 (the combination of Pointer Network and T-5 (Text-to-text transfer Transformer) algorithms) that we used in the Financial Narrative Summarisation (FNS) 2020 task. The proposed method uses Pointer networks to extract important narrative sentences from the report, and then T-5 is used to paraphrase extracted sentences into a concise yet informative sentence. We evaluate our method using $\operatorname{ROUGE}$-N (1,2), L,and SU4. The proposed method achieves the highest precision scores in all the metrics and highest F1 scores in $\operatorname{ROUGE}$ 1,and LCS and only solution to cross MUSE solution baseline in $\operatorname{ROUGE}$-LCS metrics.
\end{abstract}
\section{Introduction}
\label{intro}
Annual Reports may extend up to 250 pages long as stated above, which contains different sections General Corporate Information, financial and operating cost, CEOs message, Narrative texts, accounting policies, Financial statement including balance sheet and summary of financial data documents. In the Financial narrative summarisation task, only the narrative section is summarised, which is not explicitly marked in the dataset, making it challenging and interesting.  In recent years, previous manual small-scale research in the Accounting and Finance literature has been scaled up with the aid of NLP and ML methods, for example, to examine approaches to retrieving structured content from financial reports, and to study the causes and consequences of corporate disclosure and financial reporting outcomes ~\cite{el2018first}. \par
Companies produce glossy brochures of annual reports with a much looser structure, and this makes automatic summarisation of narratives in UK annual reports a challenging task \cite{elhaj-fns-2020}. Hence we summarize the narrative section of annual reports, particular narrative sentences that are spread loosely across the document need to be first identified and summarise those sentences. The summarisation limit is set to 1000 words, where the actual length of the report may go up to 250 pages long. Hence to summarize these long annual reports using a combination of extractive and abstractive summarisation.\par
The text summary method can be classified into two paradigms: extractive and abstractive.
The extractive summarisation method extracts the meaningful sentences or a section of text from the original text and combines them (ranked or unranked) to form a summary \cite{cheng2016neural,narayan2018ranking,yasunaga2017graph,see2017get}. Whereas abstractive summarisation generates words and sentences that are similar in meaning to the given text to form a summary that may not be in actual text \cite{nallapati2016abstractive,rush2015neural,paulus2017deep,li2018actor}. When summarizing long documents such as in our case up to 250 pages long, extractive summarisation may not produce a coherent and readable summary, and abstractive summarisation cannot cover complete information using encoder-decoder architecture. One
problem is that typical seq2seq frameworks often generate unnatural summaries consisting of repeated words or phrases \cite{li2018actor}. Hence, we come up with a combination of extractive and abstractive summarisation to first select important narrative sentences and concisely convey them. \par
Pointer Networks \cite{vinyals2015pointer} is used in various combinatorial optimization problems, such as Travelling Salesman Problem (TSP), Convex hull optimization. We used pointer networks in our task of financial narrative summarization to extract relevant narrative sentences in a particular order to have a logical flow in summary. These extracted sentences are paraphrased to summarise these sentences in an abstractive way using the T-5 sequence-to-sequence model. We train the complete model by optimizing the ROUGE-LCS evaluation metric through a reinforcement learning objective.

%
% The following footnote without marker is nebe fireded for the camera-ready
% version of the paper.
% Comment out the instructions (first text) and uncomment the 8 lines
% under "final paper" for your variant of English.
% 
\blfootnote{
    %
    % for review submission
    %
    \hspace{-0.65cm}  % space normally used by the marker
    
    %
    % % final paper: en-uk version 
    %
    % \hspace{-0.65cm}  % space normally used by the marker
    % This work is licensed under a Creative Commons 
    % Attribution 4.0 International Licence.
    % Licence details:
    % \url{http://creativecommons.org/licenses/by/4.0/}.
    % 
    % % final paper: en-us version 
    
    \hspace{-0.65cm}  % space normally used by the marker
    This work is licensed under a Creative Commons 
    Attribution 4.0 International License.
    License details:
    \url{http://creativecommons.org/licenses/by/4.0/}.
}
\section{Related Works}
\label{related_work}
In this section and Appendix \ref{ext_related_work}, we discuss related works in the fields of abstractive summarisation, extractive summarisation, combinations of these two methods, reinforcement learning applications, and summarisation of Financial Narratives and their methodology. Studies of human summarizers show that it is common to apply various operations while condensing, such as paraphrasing, generalization, sentence-level summarisation and reordering \cite{jing2002using}. We continue related work in Appendix \ref{ext_related_work}.\par
\section{Data Description}
\label{data_description}
The financial narrative summarisation dataset contains 3,863 annual reports for firms listed on LSE covering the period between 2002 and 2017 \cite{el2019multilingual,el2014detecting}. Dataset is randomly split into training (75\%), testing and validation (25\%) Table:\ref{tab:data_des}. We use NLTK sentence tokenizer \footnote{\url{https://www.nltk.org/api/nltk.tokenize.html}} to tokenize sentence in the annual report and summary processing for all our experiments. Data is further described and anayzed in Appendix \ref{data_analysis}.

\section{Methodology}
\label{mehtod}
\subsection{Model Description}
Our model is composed of three parts that are first trained or executed individually and then finally brought together using policy gradient algorithm. As stated earlier the model is combination of extractive as well as abtractive methods. Initially dataset is provided in the form \begin{math}\{x_i,y_i^j\}, 1<=i<=N, 1<=j<=7.\end{math} Here $i$ represents number of annual reports, $j$ represents number of summaries for each annual reports and $x,y$ represent report and summary respectively.\par
In extraction process we assume that for every summary sentence there is matching sentence in the annual report. To train exaction model we need these corresponding sentences in the reports. Since, annual reports are not marked explicitly with sentences we followed ROUGE scores to extract these sentences as done in \cite{chen2018fast,nallapati2016abstractive}.For every summary sentence we calculate ROUGE with every sentence in the report and then choose the sentence with maximum value. 
\begin{equation}\label{eqn_1}
j_{t}=\operatorname{argmax}_{i}\left(\text { ROUGE-L }_{\text {recall}}\left(d_{i}, s_{t}\right)\right)
\end{equation}
In equation \ref{eqn_1} $j_t$ represent sentence with maximum ROUGE score for summary sentence $s_t$ for every sentence in report $d_i$. For every report in training set there are multiple summaries. We calculate $ROUGE-L summary$ for extracted sentences as mentioned in \cite{lin2004rouge}.
\begin{equation}\label{eqn_2}
    R_{l c s}=\frac{\sum_{i=1}^{n} L C S_{U}\left(r_{i}, c_{i}\right)}{m}
\end{equation}
In equation \ref{eqn_2} $R_{l c s}$ represents ROUGE-L recall, $r_{i}, c_{i}$ are report and summary sentences respectively and $m$ is total number of words in extracted report sentences. Once summary level ROUGE-L recall is calculated we choose summary with maximum value for further processing.
Once proxy sentences and a summary is selected applying above methods extraction model is trained. In extraction model sentence level representation of report sentences is calculated using hierarchical word to sentence level Bidirectional Long Short Term Memory (Bi-LSTM)\cite{hochreiter1997long}. First Bi-LSTM is applied to word sequence of the sentence to get sentence level semantic information. Then Bi-LSTM is applied to sentences representations to get document level information in each sentence representation. We train attention mechanism \cite{bahdanau2014neural} based Pointer Networks \cite{vinyals2015pointer} different from copy mechanism used in \cite{see2017get}. Given these proxy sentences which we treat as ground truth and sentences extracted using pointer network, we train it to minimize cross-entropy loss.\par
Once sentences are extracted using above methods we fine-tune T-5 based sequence-to-sequence model for abstraction. T-5 architecture is pretrained on C-4 dataset \footnote{\url{https://www.tensorflow.org/datasets/catalog/c4}} using denoising method similar to that of Bert \cite{devlin2018bert} produce better result in language modelling. T-5 treats every task classification, summarisation, question answering, and translation as text-to-text format. These extracted sentences are taken as input and ground truth summary sentences are taken as output which is trained using cross-entropy loss. Input and output are prepared using T-5 tokenizer which outputs input\_ids and attention masks for input and target. These are then fed to model for training.\par
Once these individual components are trained individually, final complete model is trained using policy gradient algorithm with similar process as in \cite{chen2018fast}. At every extraction step agent samples an action to extract document sentence an receive reward $r(t+1)$ which is $\operatorname{ROUGE-L}_{F_{1}}$ between output from T-5 after abstraction and ground truth summary sentence.
\begin{equation}\label{reward_eqn}
r(t+1)=\operatorname{ROUGE-L}_{F_{1}}\left(abtraction\left(d_{j_{t}}\right), s_{t}\right)
\end{equation} 
The model is trained using  advantage actor-critic model to mitigate bias incurred in REINFORCE \cite{williams1992simple}.
Overall idea of our method is that first proxy sentences are extracted using ROUGE score maximisation, then extraction model is trained to extract unique narrative sentences from the report then these sentences are paraphrased using T-5 algorithm for abstraction to give concise yet informative sentence. Reinforcement learning helps to maximise ROUGE score by rewarding good sentences that are extracted and penalising bad sentences.
\begin{figure}[!ht]\label{fns}
\centering
\includegraphics[scale=0.8]{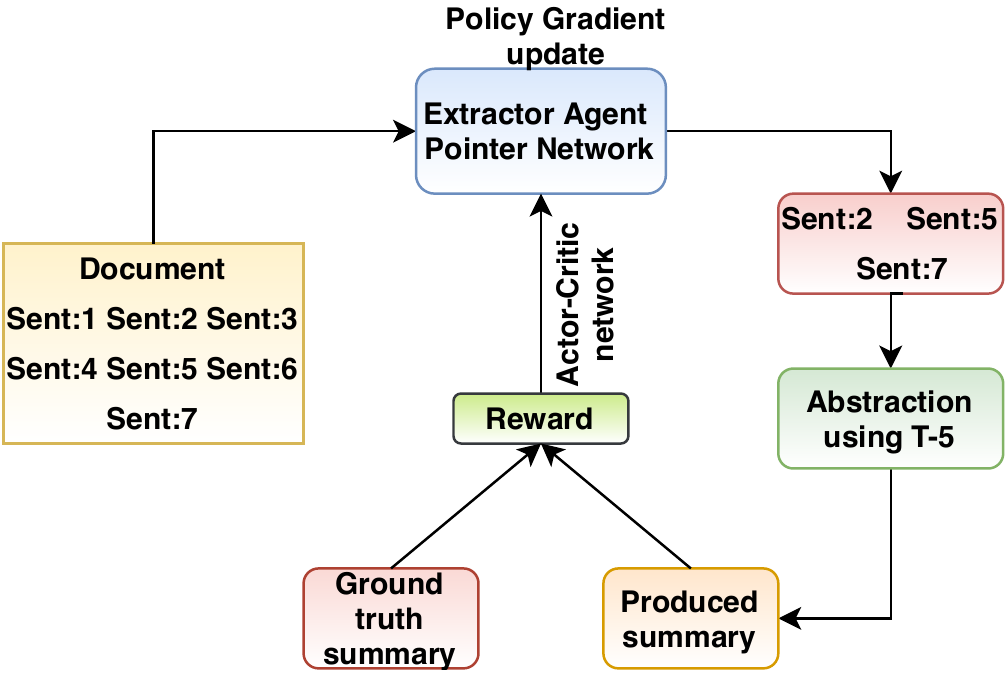}
\caption{Complete method diagram}
\end{figure}
\subsection{Parameter Tuning}
During abstraction we limit maximum number of sentences to 80, since there is limit on word limit of 1000 words and most of the reports' narrative can be summarised in less than 80 sentences as seen in section:\ref{data_description}. Word2vec \cite{mikolov2013efficient} embedding is used in representation of word in extractor model.. Vocab size is limited to $20000$, embedding size is 300, maximum number of words in a sentence 60. Model is trained using Adam optimizer with learning rate of 0.001, decay rate of 0.5. Gradient norm is clipped at $1.0$. Beam size if fixed to 2 in T-5 network, repetition penalty of 2.0. Rouge-LCS is used to optimize RL training. Abtractor and extractor networks are trained using cross-entropy loss. Traiing is done on Tesla-K80 12Gb colab GPUs with batch size of 16 and check point frequency of 16 batches. 
\section{Results and Analysis}
In this section we present results from our experiments and compare with different baselines MUSE \cite{litvak2010new}, Text-rank \cite{mihalcea2004textrank}, Lex-Rank \cite{erkan2004lexrank}, and Polynomial Summarisation \cite{litvak2013mining}. We train three models in our experiments (PoinT-5) Pointer Network with T-5, Pointer Network alone, and Pre-trained Bert for text summarisation.
\begin{table}[!ht]
\centering
\begin{tabular}{|c|c|c|c|c|c|c|c|}
\hline
\textbf{Metrics} &
  \textbf{Text-Rank} &
  \textbf{Lex-Rank} &
  \textbf{Polynomial} &
  \textbf{MUSE} &
  \textbf{\begin{tabular}[c]{@{}c@{}}Pre-trained\\ Bert\end{tabular}} &
  \textbf{\begin{tabular}[c]{@{}c@{}}Pointer\\ Net\end{tabular}} &
  \textbf{PoinT-5} \\ \hline
\textbf{Precision(R-L)}   & 0.235 & 0.210 & 0.260 & 0.470 & 0.213 & $0.603^*$ & \textbf{0.605} \\ \hline
\textbf{Recall(R-L)}      & 0.197 & 0.263 & 0.177 & 0.370 & 0.254 & 0.377 & 0.377          \\ \hline
\textbf{F-1(R-L)}         & 0.206 & 0.218 & 0.205 & 0.407 & 0.225 & $0.455^*$ & \textbf{0.456} \\ \hline
\textbf{Precision(R-1)}   & 0.414 & 0.337 & 0.324 & 0.483 & 0.241 & $0.611^*$ & \textbf{0.612} \\ \hline
\textbf{Recall (R-1)}     & 0.118 & 0.269 & 0.253 & 0.413 & 0.378 & 0.392 & 0.393          \\ \hline
\textbf{F-1 (R-1)}        & 0.172 & 0.264 & 0.274 & 0.433 & 0.283 & $0.465^*$ & \textbf{0.466} \\ \hline
\textbf{Precision (R-2)}  & 0.229 & 0.193 & 0.147 & 0.311 & 0.114 & $0.448^*$ & \textbf{0.451} \\ \hline
\textbf{Recall (R-2)}     & 0.044 & 0.107 & 0.088 & 0.198 & 0.138 & 0.220 & .0.222         \\ \hline
\textbf{F-1 (R-2}         & 0.070 & 0.120 & 0.105 & 0.234 & 0.118 & 0.289 & 0.289          \\ \hline
\textbf{Precision(R-SU4)} & 0.302 & 0.253 & 0.213 & 0.375 & 0.165 & $0.506^*$ & \textbf{0.508} \\ \hline
\textbf{Recall(R-SU4)}    & 0.048 & 0.117 & 0.105 & 0.201 & 0.149 & 0.208 & 0.209          \\ \hline
\textbf{F-1(R-SU4)}       & 0.079 & 0.140 & 0.135 & 0.253 & 0.149 & 0.286 & 0.288          \\ \hline
\end{tabular}
\caption{ROUGE Evaluation on Financial Narrative Summarisation data (Bold represents highest overall and * represents second-highest overall)}
\label{tab:rouge_evaluation}
\end{table}
In table \ref{tab:rouge_evaluation} bold is marked for highest value amongst all the solutions for the task including baselines. Pre-trained Bert for summarisation is not fine-tuned for this specific task hence its performance is not as good as Pointer Network and PoinT-5 (Pointer Network + T-5). PoinT-5 network gives highest results for precision on all the evaluation metrics ROUGE -1,2,L,SU4. From this it can be inferred that generated summmaries are highly precise in extracting narrative sentences and matches with ground truth summaries. Recall is comparatively much lower than precision in most of the evaluated metrics which means that generated summaries do not cover all the information in ground-truth summaries. This large difference in precision and recall shows that generated summaries do not cover all the sentences possibly due to restriction imposed on the number of sentences during training to follow word limit in summaries. Less recall and high precision value shows that generated summaries provide highly relevant information but does not cover complete information in ground truth summaries. From results it is evident that there is not much difference in performance between pointer net performance and PoinT-5 which extraction played key role in the architecture. PoinT-5 and Pointer Networks are the only system to cross MUSE baseline in $\operatorname{ROUGE-L}$ by atleat 5 points. These models give highest F1 results in ROUGE-L and ROUGE\_1 metrics and highest precision in ROUGE-L,1,2,SU4. 
\section{Conclusion and Future work}
In this work we present our solution on Financial Narrative Summarisation(FNS2020) dataset using PoinT-5 method explained in \ref{mehtod}. It is combination of both extractive and abstractive methods using Pointer Network and T-5. With these methods we are able to achieve highest precision score in every evaluation metric and achieve highest F-1 scores in ROUGE-LCS and ROUGE-1.\par
In our future work we would like to address several limitation of our method such as factual correctness in summaries which is very important in financial domain as done in \cite{zhang2019optimizing} in summarizing radiology reports. To improve precision of our generated summaries under 1000 words we would formulate a penalty if system generates more than 1000 words during training of RL algorithm rather than restricting algorithm to fixed number of sentences.

\bibliographystyle{coling}
\bibliography{coling2020}
\appendix
\appendixpage
\section{Data Analysis}\label{data_analysis}
\begin{figure}[!ht]

\begin{subfigure}{0.33\textwidth}
\includegraphics[scale=0.3]{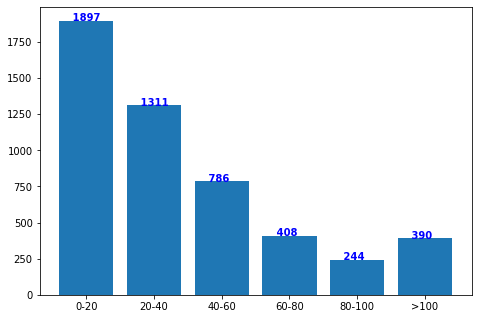} 
\caption{Summaries}
\label{fig:summary_sent}
\end{subfigure}
\begin{subfigure}{0.33\textwidth}
\includegraphics[scale=0.3]{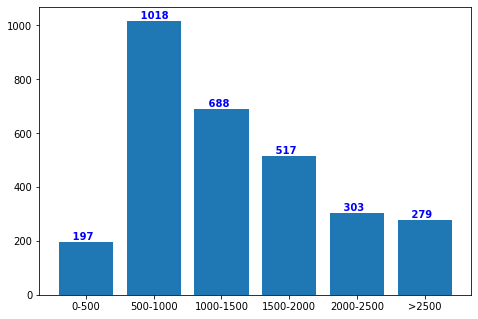}
\caption{Annual Reports}
\label{fig:report_sent}
\end{subfigure}
\begin{subfigure}{0.33\textwidth}
\includegraphics[scale=0.3]{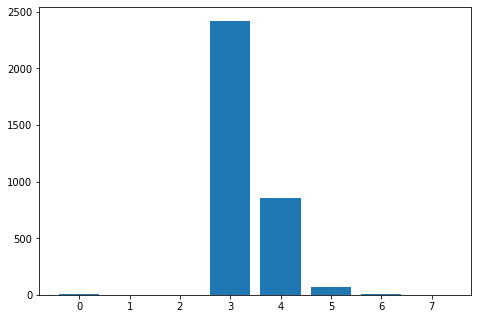}
\caption{Summary Distri}
\label{fig:summary_dist}
\end{subfigure}

\caption{Sentence wise Distribution (a,b), Number of Summary Distribution (c) }
\label{fig:sentence_distri}
\end{figure}

Table \ref{tab:data_des} presented total summaries and annual reports in train, validation, and test set. During our analysis, we found that most of the annual reports contain 100-200 sentences. There are 279 summaries with more than 500 sentences, which is large. Whereas in summaries, the average number of sentences is 50. Therefore summaries are one forth on average of annual reports and up to one-tenth in many cases. We present these analysis in Figure \ref{fig:sentence_distri}.
\begin{table}[ht]
\centering
\begin{tabular}{|c|c|c|c|}
\hline
\textbf{Data Type}        & \textbf{Training} & \textbf{Validation} & \textbf{Testing} \\ \hline
\textbf{Report full text} & 3,000             & 363                 & 500              \\ \hline
\textbf{Gold Summaries}   & 9,873             & 1,250               & 1673             \\ \hline
\end{tabular}
\caption{FNS 2020 Shared Task Dataset}
\label{tab:data_des}
\end{table}
% \begin{figure}[ht]
% \centering
% \includegraphics[scale=0.3]{summary_dist.png}
% \caption{Number of Summaries Distribution}
% \label{fig:summary_distribution}
% \end{figure}
% \begin{figure}[!ht]
% \centering
% \includegraphics[scale=0.3]{sentence_distri.png}
% \caption{Number of Sentence Distribution in annual reports}
% \label{fig:sentence_distri}
% \end{figure}
% \begin{figure}[!ht]
% \centering
% \includegraphics[scale=0.3]{summary_sentence_distri.png}
% \caption{Number of Sentence Distribution in Summaries}
% \label{fig:sentence_distri_summary}
% \end{figure}

\section{Extended Related Work}\label{ext_related_work}
\cite{li2018actor} proposes a training framework based on the actor-critic model. They apply the attention-based sequence-to-sequence model as the actor to conduct summary generation. For the critic, they combine the maximum likelihood estimator with a well designed global summary quality estimator. \cite{nallapati2016abstractive} propose RNN based encoder-decoder model for abstractive summarisation. They apply bi-directional GRU-RNN at the encoder side and uni-directional GRU in decoder with attention. In their approach, each mini-batch's decoder-vocabulary is restricted to words in the source documents of that batch. \cite{rush2015neural} also propose attention based sequence to sequence model for abstractive summarisation. \cite{paulus2017deep} states that attentional, RNN-based encoder-decoder models for abstractive summarisation have achieved good performance on short input and output sequences. For longer documents and summaries, however, these models often include repetitive and incoherent phrases. Hence they combine standard word prediction with the global sequence prediction training of RL which makes resulting summaries become more readable.\par
\cite{narayan2018ranking} propose a reinforcement learning-based sentence ranking approach in extractive summarisation. \cite{cheng2016neural} use attention architecture for words and sentence level extraction in extractive summarisation method. \cite{yasunaga2017graph} proposes a multi-document summarization system that exploits the representational power of deep neural networks and the sentence relation information encoded in graph representations of document clusters. Specifically, they apply Graph Convolutional Networks on sentence relation graphs.\par
\cite{see2017get} propose a novel \say{Pointer Generator} networks for abstractive summarisation. In this model a word is chosen with probability \textit{P\textsubscript{gen}} from overall vocabulary and \textit{$1-P\textsubscript{gen}$} from current sentence using attention weights. They apply coverage mechanism by \cite{tu2016modeling} to avoid word repetitions in the summary which is common in large documents.\par
\cite{chen2018fast} apply combination abstractive and extractive summarisation by join training using reinforcement learning. They apply pointer network for extraction (different from Pointer Generator) and RNN based encoder-decoder for abstraction. \cite{hsu2018unified} propose unified extractive and abstractive with a hierarchical sentence and word level attention model using novel inconsistency loss. \cite{wang2019text} used T-5 \cite{raffel2019exploring} based sentence representation for combined extractive and abstractive training.\par
Financial Narrative Summarisation has been explored in the past by \cite{cardinaels2018automatic}. They provide multiple evidence that algorithm-based summaries are less positively biased than management summaries.

\end{document}